\documentclass[11pt]{article}

\usepackage[preprint]{acl}

\usepackage{times}
\usepackage{latexsym}
\usepackage{booktabs}
\usepackage[T1]{fontenc}
\usepackage{float}
\usepackage[utf8]{inputenc}

\usepackage{amsfonts}

\usepackage{microtype}

\usepackage{inconsolata}

\usepackage{graphicx}
\author{
Mahmoud Salhab \\
CNTXT AI \\
Abu Dhabi, UAE \\
\texttt{mahmoud.salhab@cntxt.tech}
\And
Shameed Sait \\
CNTXT AI \\
Abu Dhabi, UAE \\
\texttt{shameed.ali@cntxt.tech}
\AND
Mohammad Abusheikh \\
CNTXT AI \\
Abu Dhabi, UAE \\
\texttt{mas@cntxt.tech}
\And
Hasan Abusheikh \\
CNTXT AI \\
Abu Dhabi, UAE \\
\texttt{has@cntxt.tech}
}

\title{Munsit at NADI 2025 Shared Task 2: Pushing the Boundaries of Multidialectal Arabic ASR with Weakly Supervised Pretraining and Continual Supervised Fine-tuning}

\begin{document}
\maketitle

\begin{abstract}
Automatic speech recognition (ASR) plays a vital role in enabling natural human–machine interaction across applications such as virtual assistants, industrial automation, customer support, and real-time transcription. However, developing accurate ASR systems for low-resource languages like Arabic remains a significant challenge due to limited labeled data and the linguistic complexity introduced by diverse dialects. In this work, we present a scalable training pipeline that combines weakly supervised learning with supervised fine-tuning to develop a robust Arabic ASR model. In the first stage, we pretrain the model on 15,000 hours of weakly labeled speech covering both Modern Standard Arabic (MSA) and various Dialectal Arabic (DA) variants. In the subsequent stage, we perform continual supervised fine-tuning using a mixture of filtered weakly labeled data and a small, high-quality annotated dataset. Our approach achieves state-of-the-art results, ranking first in the multi-dialectal Arabic ASR challenge. These findings highlight the effectiveness of weak supervision paired with fine-tuning in overcoming data scarcity and delivering high-quality ASR for low-resource, dialect-rich languages.

\end{abstract}

\section{Introduction}

Automatic speech recognition (ASR), or speech-to-text (STT), converts spoken language into text, enabling voice-based interaction with machines \cite{10.1007/978-3-030-21902-4_2, kheddar2024automatic}. ASR is widely applied in healthcare, robotics, law enforcement, telecommunications, smart homes, and consumer electronics, among other domains \cite{vajpai2016industrial}. Arabic, the fourth most used language online and one of the UN’s six official languages, remains underrepresented in ASR research despite serving millions across 22 countries \cite{6841973}.

Arabic exists in three forms: Classical Arabic (CA), the language of historical and religious texts; Modern Standard Arabic (MSA), used in formal contexts; and Dialectal Arabic (DA), comprising diverse regional variants \cite{ALAYYOUB2018522}. While some datasets, such as MASC \cite{e1qb-jv46-21} and SADA \cite{10446243}, have advanced Arabic ASR, they remain limited in size and linguistic diversity, hindering model generalization. Neural ASR systems require vast transcribed datasets \cite{lu2020exploring, 9414087}, but manual transcription is costly and time-intensive \cite{gao2023learning}.

We address this by proposing a weakly supervised Arabic ASR system based on the Conformer architecture \cite{gulati2020conformerconvolutionaugmentedtransformerspeech}, trained on large-scale weakly labeled MSA and DA speech. In the first stage, we pretrain the model on 15,000 hours of weakly labeled speech covering both Modern Standard Arabic (MSA) and various Dialectal Arabic (DA) variants. In the subsequent stage, we perform continual supervised fine-tuning using a mixture of filtered weakly labeled data and a small, high-quality annotated dataset. This approach eliminates the need for extensive manual transcription and achieves state-of-the-art results on standard benchmarks, demonstrating the potential of weak supervision for low-resource languages.

\section{Background}
\label{sec:related-work}
Arabic Automatic Speech Recognition (ASR) remains challenging due to data scarcity, lexical variation, morphological complexity, and dialect diversity across 22 Arab countries \cite{ali2014advances, Cardinal2014, diehl2012morphological}. Traditional systems often used hybrid HMM-DNN pipelines \cite{Cardinal2014, bouchakour2018improving}.

Dialectal variation is a major bottleneck, as most systems focus on Modern Standard Arabic (MSA) and high-resource dialects, performing poorly on low-resource varieties \cite{djanibekov-etal-2025-dialectal}. To address this, \citeauthor{djanibekov-etal-2025-dialectal} released open-source ASR models covering 17 countries, 11 dialects, and code-switched Arabic-English/French speech. Other efforts integrate dialect identification directly into ASR \cite{waheed2023voxarabica} or build dialect-specific systems, e.g., for Egyptian \cite{10.1109/ICASSP.2013.6639311} and Algerian Arabic \cite{MENACER201781}.

End-to-end architectures have advanced Arabic ASR by eliminating the need for intermediate feature extraction \cite{radford2023robust}. Notable examples include large-scale weakly supervised systems such as Whisper \cite{pmlr-v202-radford23a}. Weak supervision has proven particularly effective; for instance, \cite{salhab2025advancingarabicspeechrecognition} trained a Conformer model from scratch on 15,000 hours of weakly labeled MSA and dialectal speech, achieving state-of-the-art results without relying on manual transcription.

\section{Methodology}
\label{sec:methodology}

Our approach consists of two main stages: weakly supervised pretraining followed by continual supervised fine-tuning. In the first stage, we train the model on a large-scale, diverse speech dataset with weak labels—labels that are not guaranteed to be accurate (i.e., not manually verified)—in line with the strategy proposed in \cite{salhab2025advancingarabicspeechrecognition}.

In the second stage, the pretrained model is further fine-tuned using a smaller, high-quality dataset constructed from two main sources: (1) the official training data released for the task (the Casablanca training set \cite{talafha2024casablancadatamodelsmultidialectal}), which is expanded through various augmentation techniques; and (2) a filtered subset derived from the initial 15,000 hours of weakly labeled training data, selected through a rigorous data cleaning and filtering process.

\begin{figure*}[t]
  \centering
  \includegraphics[width=0.75\textwidth]{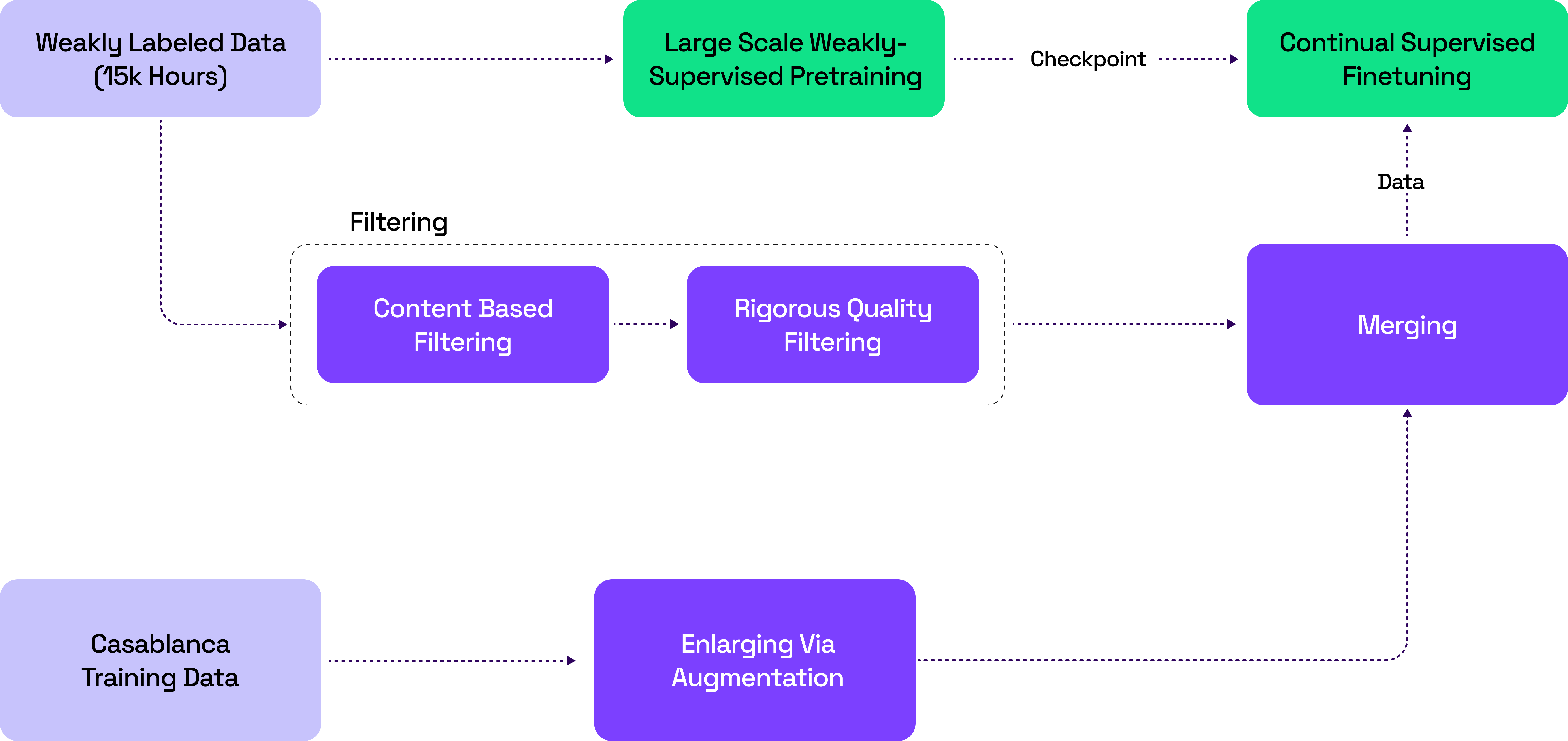}
  \caption{The solution's full pipeline encompasses large-scale pretraining followed by continual fine-tuning.}
  \label{fig:solution}
\end{figure*}

An overview of the complete pipeline is presented in Figure~\ref{fig:solution}. The following subsections provide a detailed explanation of each stage of the proposed approach.

\subsection{Weakly Supervised Learning}

Traditional supervised ASR training uses high-quality, human-annotated pairs \((x_i, y_i)\), where the input \(x_i\) is typically a mel-spectrogram and the output \(y_i\) consists of a sequence of tokens, each selected from a predefined vocabulary. These accurate labels are assumed to be independently drawn from a clean data distribution, enabling the model to learn a function that performs well on unseen test examples.

On the other hand, weakly supervised learning depends on automatically generated or crowd-sourced labels \(\widehat{y}_i\), which may contain errors or noise. These weak labels come from a noisier distribution and might not precisely reflect the true transcription. Nonetheless, models trained on such data aim to generalize effectively when evaluated on clean datasets.

Building upon the approach introduced in \cite{salhab2025advancingarabicspeechrecognition}, we adopted the same training pipeline and experimental settings to develop the initial foundation model. Specifically, the model was trained on 15,000 hours of weakly annotated speech data, with automatic labeling performed using the same method described in the aforementioned work.

\subsection{Continual supervised finetuning}

In neural network-based ASR systems, training typically begins either from scratch—with randomly initialized weights and a large training corpus—or from a pretrained model that has already been exposed to a large-scale dataset. The latter approach enables faster convergence and often better generalization on the target task due to prior knowledge encoded in the pretrained weights.

In this stage, we adopt the second strategy by initializing the model with weights obtained from the first stage, which was trained on weakly labeled data. We then fine-tune this model using a smaller yet higher-quality dataset comprising 3,000 hours of filtered weakly annotated data. The filtering process was designed to exclude news content—largely composed of Modern Standard Arabic (MSA)—and to retain only segments that passed stringent quality thresholds, as outlined in the pipeline of \cite{salhab2025advancingarabicspeechrecognition}. Additionally, we incorporate the Casablanca Challenge training dataset, which is further expanded through various data augmentation techniques. Unlike the first stage that relied on noisy supervision, this fine-tuning phase leverages only high-quality transcriptions.

\subsection{Model Architecture}

The Conformer architecture \cite{gulati2020conformerconvolutionaugmentedtransformerspeech} effectively models both long- and short-range dependencies in speech through a combination of convolutional modules and multi-head self-attention, making it highly suitable for automatic speech recognition. In this work, we adopt the same architecture as introduced in the original paper, specifically using the large variant of the model.

\subsection{Experimental Setup}
\label{sec:experiments}

Our ASR experiments utilized the Conformer architecture trained with the Connectionist Temporal Classification (CTC) objective. To tokenize the transcripts, we employed a SentencePiece model trained on the same training corpus, with a vocabulary of 128 tokens.

Model training was carried out in a distributed setting across 8 NVIDIA A100 GPUs using a global batch size of 512. Input features were 80-dimensional mel-spectrograms, extracted using a 25~ms frame length and a 10~ms hop size.

During the weakly supervised pretraining phase, optimization was performed using the AdamW optimizer combined with the Noam learning rate schedule, incorporating 10{,}000 warm-up steps and peaking at a learning rate of $2 \times 10^{-3}$. For regularization purposes, we applied a dropout rate of 0.1 across all layers and used L2 weight decay. For the fine-tuning stage, the learning rate was reduced by a factor of ten.

To optimize training speed and reduce memory overhead, computations were performed using \texttt{bfloat16} precision. The Conformer model was initialized with random weights and comprised 18 encoder layers. Each layer featured a hidden dimension of 512, 8 attention heads, a convolutional kernel size of 31, and a feedforward expansion factor of 4. The complete model architecture contained approximately 121 million parameters.

\subsection{Evaluation Metrics \& Datasets}


The model’s performance was evaluated using Word Error Rate (WER) and Character Error Rate (CER). Training used a development set with paired speech and transcriptions, while testing involved blind evaluation on speech-only data via CodeBench.

\section{Results}
\label{sec:conclusion}

\begin{table*}[ht]
\centering
\caption{Dialect-wise WER (\%) Comparison Across Participants.}
\label{tab:wer}
\begin{tabular}{lccccccccc}
\toprule
\textbf{Participant} & \textbf{Avg} & \textbf{JOR} & \textbf{EGY} & \textbf{MOR} & \textbf{ALG} & \textbf{YEM} & \textbf{MAU} & \textbf{UAE} & \textbf{PAL} \\
\midrule
\textbf{msalhab96 (Ours)} & \textbf{35.69} & 20.68 & 20.89 & 41.72 & 53.62 & 44.62 & 59.03 & 22.67 & 22.28 \\
youssef\_saidi & 38.54 & 28.03 & 26.83 & 38.27 & 53.73 & 46.63 & 58.11 & 29.35 & 27.36 \\
yusser & 39.78 & 28.84 & 29.50 & 43.07 & 55.04 & 46.42 & 59.37 & 28.38 & 27.66 \\
alhassan10ehab & 42.05 & 32.25 & 24.73 & 48.22 & 60.32 & 51.77 & 66.23 & 28.01 & 24.87 \\
badr\_alabsi & 44.15 & 31.74 & 37.24 & 43.31 & 56.12 & 46.15 & 63.32 & 38.65 & 36.63 \\
Baseline & 93.90 & 46.10 & 100.07 & 100.38 & 101.03 & 101.09 & 100.59 & 101.15 & 100.77 \\
rafiulbiswas & 104.90 & 44.97 & 113.98 & 104.08 & 116.60 & 113.54 & 111.59 & 116.79 & 117.61 \\
\bottomrule
\end{tabular}
\end{table*}

\begin{table*}[ht]
\centering
\caption{Dialect-wise CER (\%) Comparison Across Participants.}
\label{tab:cer}
\begin{tabular}{lccccccccc}
\toprule
\textbf{Participant} & \textbf{Avg} & \textbf{JOR} & \textbf{EGY} & \textbf{MOR} & \textbf{ALG} & \textbf{YEM} & \textbf{MAU} & \textbf{UAE} & \textbf{PAL} \\
\midrule
\textbf{msalhab96 (Ours)} & \textbf{12.21} & 5.64 & 7.33 & 14.04 & 18.44 & 14.30 & 23.28 & 6.55 & 8.06 \\
youssef\_saidi & 14.53 & 9.36 & 11.44 & 13.66 & 20.43 & 16.66 & 24.53 & 9.91 & 10.20 \\
yusser & 14.76 & 9.47 & 11.91 & 15.52 & 20.59 & 16.05 & 24.85 & 9.04 & 10.59 \\
alhassan10ehab & 16.19 & 9.90 & 10.21 & 18.12 & 23.34 & 20.41 & 29.11 & 8.99 & 9.41 \\
badr\_alabsi & 15.59 & 9.95 & 12.57 & 15.07 & 21.39 & 15.69 & 26.70 & 11.15 & 12.19 \\
Baseline & 72.79 & 19.29 & 81.38 & 80.42 & 79.59 & 80.58 & 82.89 & 80.28 & 77.93 \\
rafiulbiswas & 84.69 & 19.19 & 97.66 & 87.59 & 94.27 & 94.56 & 92.85 & 97.01 & 94.42 \\
\bottomrule
\end{tabular}
\end{table*}

\begin{table}[ht]
\centering
\caption{Comparison of WER (\%) Across Evaluation and Testing Datasets.}
\label{tab:wer_eval_test}
\begin{tabular}{lcc}
\toprule
\textbf{Dialect} & \textbf{Evaluation} & \textbf{Testing} \\
\midrule
Avg  & 36.83 & 35.69 \\
JOR  & 21.52 & 20.68 \\
EGY  & 22.89 & 20.89 \\
MOR  & 44.20 & 41.72 \\
ALG  & 54.78 & 53.62 \\
YEM  & 47.69 & 44.62 \\
MAU  & 57.62 & 59.03 \\
UAE  & 24.05 & 22.67 \\
PAL  & 21.91 & 22.28 \\
\bottomrule
\end{tabular}
\end{table}


\begin{table}[ht]
\centering
\caption{Comparison of CER (\%) Across Evaluation and Testing Datasets.}
\label{tab:cer_eval_test}
\begin{tabular}{lcc}
\toprule
\textbf{Dialect} & \textbf{Evaluation} & \textbf{Testing} \\
\midrule
Avg  & 11.94 & 12.21 \\
JOR  & 5.39  & 5.64  \\
EGY  & 7.50  & 7.33  \\
MOR  & 14.06 & 14.04 \\
ALG  & 17.71 & 18.44 \\
YEM  & 14.73 & 14.30 \\
MAU  & 21.73 & 23.28 \\
UAE  & 6.97  & 6.55  \\
PAL  & 7.40  & 8.06  \\
\bottomrule
\end{tabular}
\end{table}

We evaluate our proposed system, against all participating teams using both Word Error Rate (WER) and Character Error Rate (CER) metrics, reported across multiple Arabic dialects. The results demonstrate the robustness of our approach across both evaluation and testing phases, as well as its ability to generalize across diverse dialectal variations.

As shown in Table~\ref{tab:wer}, our system achieved the lowest average WER (35.69\%), outperforming all other submissions. Notably, our work consistently maintained lower WER in most of the dialects, particularly excelling in Jordanian (20.68\%), Egyptian (20.89\%), and Emirati (22.67\%) dialects. Similarly, Table~\ref{tab:cer} shows that our model achieved the lowest average CER (12.21\%), with the best performance observed in Jordanian (5.64\%) and Egyptian (7.33\%) dialects.
Tables~\ref{tab:wer_eval_test} and~\ref{tab:cer_eval_test} present a breakdown of WER and CER across evaluation and testing phases/datasets. The average WER decreased from 36.83\% during evaluation to 35.69\% in testing, suggesting that our model generalizes well to unseen data. This trend is consistent across most dialects. For instance, the WER in the Jordanian dialect dropped from 21.52\% to 20.68\%, and in the Yemeni dialect from 47.69\% to 44.62\%.

Similarly, the average CER exhibited a slight increase from 11.94\% (evaluation) to 12.21\% (testing), though the variation across dialects remained minimal, underscoring the model’s stability. These consistent results across both phases affirm the robustness and dialectal adaptability of our ASR system.

\section{Conclusion}
We present a scalable two-stage pipeline—pretraining on 15,000 hours of weakly labeled audio, then fine-tuning on a filtered 3,000-hour weak subset plus an augmented official training set—that, with data filtering, augmentation, and a Conformer backbone, achieved state-of-the-art performance and first place in the multi-dialectal Arabic ASR challenge, demonstrating that carefully curated weak supervision combined with targeted fine-tuning can overcome data scarcity and dialectal diversity.

\bibliography{custom}

\appendix

\end{document}